\pdfoutput=1
\documentclass[11pt]{article}
\usepackage{authblk}
\usepackage{EMNLP2022}
\usepackage{times}
\usepackage{latexsym}
\usepackage{booktabs}

\usepackage[T1]{fontenc}
\usepackage[utf8]{inputenc}
\usepackage{CJKutf8}

\usepackage{color}
\usepackage{multirow}

\usepackage{microtype}
\usepackage{inconsolata}

\usepackage{graphicx}
\graphicspath{{./figures/}} 
\usepackage{bm}
\usepackage{amsfonts}
\usepackage{amsmath}
\usepackage{amssymb}

\usepackage{makecell}
\title{\mbox{Improving Chinese Spelling Check by Character Pronunciation Prediction:}\\The Effects of Adaptivity and Granularity}
\author{Jiahao Li$^1$, Quan Wang$^2$\thanks{\hspace{0.15cm}Corresponding author: Quan Wang.} , Zhendong Mao$^1$, Junbo Guo$^3$, Yanyan Yang$^4$,  Yongdong Zhang$^1$\\ \\
  $^1$University of Science and Technology of China, Hefei, China \\
  $^2$MOE Key Laboratory of Trustworthy Distributed Computing and Service, \authorcr{Beijing University of Posts and Telecommunications, Beijing, China} \\
  $^3$People's Daily Online Co., Beijing, China \\
  $^4$People’s Public Security University of China, Beijing, China \\
  {\tt jiahao66@mail.ustc.edu.cn, wangquan@bupt.edu.cn, zdmao@ustc.edu.cn} \\
  {\tt guojunbo@people.cn, zhyd73@ustc.edu.cn}}

\begin{document}
\begin{CJK}{UTF8}{gkai}
\maketitle
\begin{abstract}

Chinese spelling check (CSC) is a fundamental NLP task that detects and corrects spelling errors in Chinese texts. As most of these spelling errors are caused by phonetic similarity, effectively modeling the pronunciation of Chinese characters is a key factor for CSC. In this paper, we consider introducing an auxiliary task of Chinese pronunciation prediction (CPP) to improve CSC, and, for the first time, systematically discuss the adaptivity and granularity of this auxiliary task. We propose SCOPE which builds on top of a shared encoder two parallel decoders, one for the primary CSC task and the other for a fine-grained auxiliary CPP task, with a novel adaptive weighting scheme to balance the two tasks. In addition, we design a delicate iterative correction strategy for further improvements during inference. Empirical evaluation shows that SCOPE achieves new state-of-the-art on three CSC benchmarks, demonstrating the effectiveness and superiority of the auxiliary CPP task. Comprehensive ablation studies further verify the positive effects of adaptivity and granularity of the task. Code and data used in this paper are publicly available at \url{https://github.com/jiahaozhenbang/SCOPE}. 

\end{abstract}

\section{Introduction}

Chinese Spelling Check (CSC), which aims to detect and correct spelling errors in Chinese texts, is a fundamental task in Chinese natural language processing. Spelling errors mainly originate from human writing errors and machine recognition errors, {\it e.g.}, errors caused by automatic speech recognition (ASR) and optical character recognition (OCR) systems \cite{DBLP:conf/acl/HuangLJZCWX20}. With the latest development of deep neural networks, neural CSC methods, in particular those based on encoder-decoder architectures, have become the mainstream of research in recent years \citep{DBLP:conf/acl/XuLZLWCHM21,DBLP:conf/acl/LiuYYZW20}. Encoder-decoder models regard CSC as a special sequence-to-sequence (Seq2Seq) problem, where a sentence with spelling errors is given as the input and a corrected sentence of the same length will be generated as the output.

\begin{table}
\small
\centering\setlength{\tabcolsep}{1pt}
\begin{tabular*}{0.48 \textwidth}{@{\extracolsep{\fill}}llcc}
\toprule
\multicolumn{2}{@{}l}{\multirow{2}{*}{Instance}} & \multicolumn{2}{c@{}}{Similarity} \\
\cmidrule(r){3-4}
& & Coarse & Fine \\
\midrule
W: & $\!$ 我觉得你们会好好的\textcolor{red}{完(wan2/w,an,2)}。  & \multirow{4}{*}{$1$} & \multirow{4}{*}{$1$} \\
                    & $\!$ I think you will finish well.  & & \\
R:  & $\!$ 我觉得你们会好好的\textcolor{blue}{玩(wan2/w,an,2)}。 & & \\
                    & $\!$ I think you will play well. & & \\
\midrule
W: & $\!$我以前想要\textcolor{red}{高(gao1/g,ao,1)}诉你。 & \multirow{4}{*}{$0$} & \multirow{4}{*}{$^2\!/\!_3$} \\
                    & $\!$I tried to high you before. & & \\
R:  & $\!$我以前想要\textcolor{blue}{告(gao4/g,ao,4)}诉你。 & & \\
                        & $\!$I tried to tell you before. & & \\
\midrule
W: & $\!$他\textcolor{red}{收(shou1/sh,ou,1)}到山上的时候。 & \multirow{4}{*}{$0$} & \multirow{4}{*}{$^1\!/\!_3$} \\
                    & $\!$When he received the mountain. & & \\
R:  & $\!$他\textcolor{blue}{走(zou3/z,ou,3)}到山上的时候。 & & \\
                        & $\!$When he walked up the mountain. & & \\
\midrule
W: & $\!$行为都被\textcolor{red}{蓝(lan2/l,an,2)}控设备录影。 & \multirow{4}{*}{$0$} & \multirow{4}{*}{$0$} \\
                        & $\!$Actions are recorded by blue control devices. & & \\
R:  & $\!$行为都被\textcolor{blue}{监(jian1/j,ian,1)}控设备录影。 & & \\
                        & $\!$Actions are recorded by surveillance devices. & & \\
\bottomrule
\end{tabular*}
\caption{\label{tab:1}Instances from SIGHAN15 \citep{DBLP:conf/acl-sighan/TsengLCC15}. For each instance, coarse-/fine-grained pinyin of the misspelled (red) and correct (blue) characters are provided, along with their phonological similarity degree (the fraction of identical components) in terms of these two types of pinyin.}
\end{table}

Previous research has shown that about 76\% of Chinese spelling errors are induced by phonological similarity \citep{DBLP:journals/talip/LiuLTCWL11}. Hence, it is a crucial factor to effectively model the pronunciation of Chinese characters for the CSC task. In fact, almost all current advanced CSC approaches have actually exploited, either explicitly or implicitly, character pronunciation. The implicit use takes into account phonological similarities between pairs of characters, {\it e.g.}, by increasing the decoding probability of characters with similar pronunciation \citep{DBLP:conf/acl/ChengXCJWWCQ20} or integrating such similarities into the encoding process via graph convolutional networks (GCNs) \citep{DBLP:conf/acl/ChengXCJWWCQ20}. The explicit use considers directly the pronunciation, or more specifically, pinyin\footnote{Pinyin is the official phonetic system of Mandarin Chinese, which literally means ``spelled sounds''.}, of individual characters, encoding the pinyin of input characters to produce extra phonetic features \citep{DBLP:conf/acl/XuLZLWCHM21,DBLP:conf/acl/HuangLJZCWX20} or decoding the pinyin of target correct characters to serve as an auxiliary prediction task \citep{DBLP:conf/acl/LiuYYZW20,DBLP:conf/emnlp/JiYQ21}.

This paper also considers improving CSC with auxiliary character pronunciation prediction (CPP), but focuses specifically on the {\it adaptivity} and {\it granularity} of the auxiliary task, which have never been systematically studied before. First, all the prior attempts in similar spirit simply assigned a universal trade-off between the primary and auxiliary tasks for all instances during training, while ignoring the fact that the auxiliary task might provide different levels of benefits given different instances. Take for example the instances shown in Table~\ref{tab:1}. Compared to the misspelled character ``蓝'' and its correction ``监'' in the 4th instance, the two characters ``完'' and ``玩'' in the 1st instance are much more similar in pronunciation, suggesting that the spelling error there is more likely to be caused by phonological similarity, to which the pronunciation-related auxiliary task might provide greater benefits and hence should be assigned a larger weight. Second, prior efforts mainly explored predicting the whole pinyin of a character, {\it e.g.}, ``gao1'' for ``高''. Nevertheless, a syllable in Chinese is inherently composed of an initial, a final, and a tone, {\it e.g.}, ``g'', ``ao'', and ``1'' for ``高''. This fine-grained phonetic representation can better reflect not only the intrinsic regularities of Chinese pronunciation, but also the phonological similarities between Chinese characters. Consider for example the ``高'' and ``告'' case from the 2nd instance in Table~\ref{tab:1}. These two characters show no similarity in terms of their whole pinyin, but actually they share the same initial and final, differing solely in their tones. 

\begin{figure*}[t]
\centering
\includegraphics[scale=0.5]{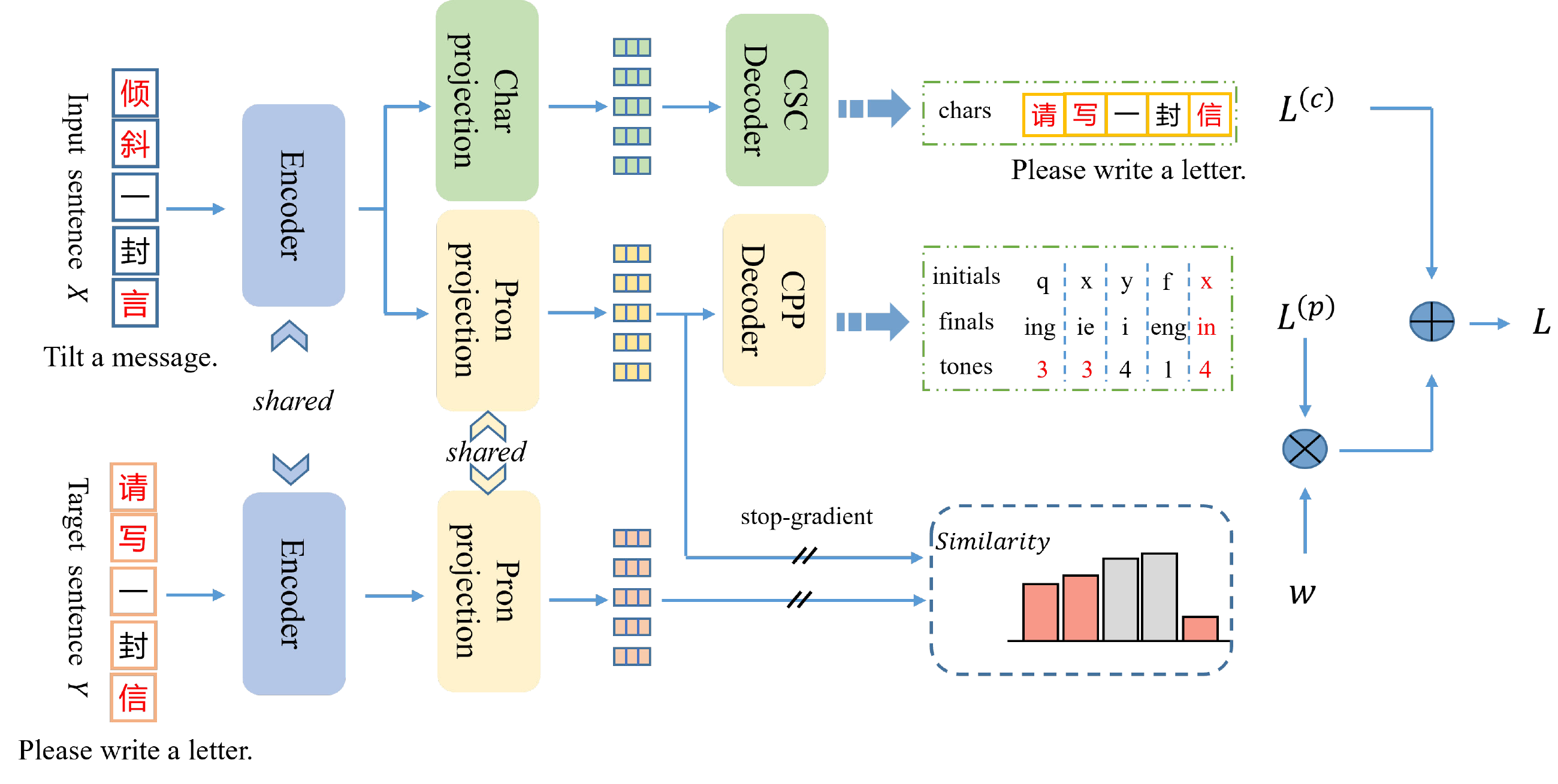}
\caption{Overview of SCOPE. \textbf{Top:} The one-encoder-two-decoder structure for CSC and CPP. The input sentence $X$ is fed into the encoder and then, after character-/pronunciation-specific  feature projection, two parallel decoders, one to predict the characters, the other to predict the initial, final, and tone of each character in the target sentence. \textbf{Bottom:} Adaptive task weighting between CSC and CPP (detached in the backward pass). The target sentence $Y$ is fed into the encoder and the pronunciation-specific feature projection layer. Then the similarities between input and target sentences on character level are calculated and the adaptive weights are accordingly defined. \textbf{Note:} Only the CSC decoder branch (along with the encoder) will be used at inference time.}
\label{fig:1}
\end{figure*}

Based on the above intuitions we devise \textbf{SCOPE} ({\it i.e.}, \textbf{S}pelling \textbf{C}heck by pr\textbf{O}nunciation \textbf{P}r\textbf{E}diction), which introduces a fine-grained CPP task with an adaptive task weighting scheme to improve CSC. Figure~\ref{fig:1} provides an overview of SCOPE. Given a sentence with spelling errors as input, we encode it using ChineseBERT \citep{DBLP:conf/acl/SunLSMAHWL20} to produce semantic and phonetic features. Then we build on top of the encoder two parallel decoders, one to generate target correct characters, {\it i.e.}, the primary CSC task, and the other to predict the initial, final and tone of the pinyin of each target character, {\it i.e.}, the auxiliary fine-grained CPP task. The trade-off between the two tasks can be further adjusted adaptively for each instance, according to the phonological similarity between input and target characters therein. In addition, we design an iterative correction strategy during inference to address the over-correction issue and tackle difficult instances with consecutive errors.

We empirically evaluate SCOPE on three shared benchmarks, and achieve substantial and consistent improvements over previous state-of-the-art on all three benchmarks, demonstrating the effectiveness and superiority of our auxiliary CPP task. Comprehensive ablation studies further verify the positive effects of adaptivity and granularity of the task.

The main contributions of this paper are summarized as follows: (1) We investigate the possibility of introducing an auxiliary CPP task to improve CSC and, for the first time, systematically discuss the adaptivity and granularity of this auxiliary task. (2) We propose SCOPE, which builds two parallel decoders upon a shared encoder for CSC and CPP, with a novel adaptive weighting scheme to balance the two tasks. (3) We establish new state-of-the-art on three benchmarking CSC datasets.

\section{Related Work}
CSC is a fundamental NLP task that has received wide attention over the past decades. Early work on this topic was mainly based on manually designed rules \citep{DBLP:conf/icml/ManguB97, jiang2012rule}. After that, statistical language models became the mainstream for CSC \citep{DBLP:conf/acl-sighan/ChenLLWC13, DBLP:conf/acl-sighan/YuL14, DBLP:conf/acl-sighan/TsengLCC15}. Methods of this kind in general followed a pipeline of error detection, candidate generation, and candidate selection. Given a sentence, the error positions are first detected by the perplexity of a language model. The candidates for corrections can then be generated according to similarity between characters, typically by using a confusion set. And the final corrections can be determined by scoring the sentence replaced by the candidates with the language model \citep{DBLP:conf/acl-sighan/LiuCLDM13, DBLP:conf/acl-sighan/XieHZHHCH15}. 

In the era of deep learning, especially after Transformer \citep{DBLP:conf/nips/VaswaniSPUJGKP17} and pre-trained language models like BERT \citep{DBLP:conf/naacl/DevlinCLT19} were proposed, a large number of neural CSC methods have emerged. \citet{DBLP:conf/aclnut/HongYHLL19} used Transformer as an encoder to produce candidates and designed a confidence-similarity decoder to filter these candidates. \citet{DBLP:conf/acl/ZhangHLL20} designed a detection network based on Bi-GRU to predict the error probability of each character and passed the probabilities to a BERT-based correction network via a soft masking mechanism. \citet{DBLP:conf/acl/ChengXCJWWCQ20} employed GCNs combined with BERT to further model interdependences between characters. Recent work of \citep{DBLP:conf/acl/XuLZLWCHM21, DBLP:conf/acl/LiuYYZW20, DBLP:conf/acl/HuangLJZCWX20} proposed to encode phonetic and glyph information in addition to semantic information, and then combine phonetic, glyph and semantic features to make final predictions. 

As we could see, modeling pronunciation information is prevailing in CSC research \citep{DBLP:conf/acl/ZhangPZWHSWW21}, typically via an encoding process to extract phonetic features. \citet{DBLP:conf/acl/LiuYYZW20} proposed the first work that considered predicting the pronunciation of target characters as an auxiliary task. Their work, however, employed pronunciation prediction in a coarse-grained, non-adaptive manner, which is quite different to ours.

\section{Our Approach}
This section presents our approach SCOPE for the CSC task. Below, we first define the problem formulation and then describe our approach in detail.

\subsection{Problem Formulation}\label{section:3.1}
The Chinese spelling check (CSC) task is to detect and correct spelling errors in Chinese texts. Given a misspelled sentence $X = \{x_1, x_2, \cdots, x_n\}$ with $n$ characters, a CSC model takes $X$ as input, detects potential spelling errors on character level, and outputs a corresponding correct sentence $Y = \{y_1, y_2,$ $\cdots,  y_n \} $ of equal length. This task can be viewed as a conditional sequence generation problem that models the probability of $p(Y|X)$. We are further given the fine-grained pinyin of each character $y_i$ in the correct sentence $Y$, represented as a triplet in the form of $(\alpha_i, \beta_i, \gamma_i)$, where $\alpha_i$, $\beta_i$, and $\gamma_i$ indicate the initial, final, and tone, respectively. Note that such kind of pinyin of the output sentence is required and provided solely during training.\footnote{In fact, we also use the pinyin of each character $x_i$ in the input sentence $X$ during the ChineseBERT encoding process (detailed later), and this kind of pinyin of the input sentence is required and provided during both training and inference.}

\subsection{SCOPE Architecture}\label{section:3.2}
The key idea of SCOPE is to employ a fine-grained character pronunciation prediction (CPP) task with an adaptive task weighting scheme to improve CSC. In achieving this SCOPE builds upon a shared encoder two parallel decoders, one for the primary CSC task and the other for the auxiliary CPP task. The trade-off between the two tasks is further determined adaptively based on the phonological similarity between input and target characters. Figure~\ref{fig:1} summarizes the overall architecture of SCOPE.

\paragraph{Encoder}
Similar to recent CSC approaches that leverage multimodal information \citep{DBLP:conf/acl/LiuYYZW20,DBLP:conf/acl/XuLZLWCHM21}, we use ChineseBERT \citep{DBLP:conf/acl/SunLSMAHWL20} as the encoder to extract semantic, phonetic, and morphologic features as well for the CSC task. ChineseBERT is a pre-trained language model that incorporates both the pinyin and glyph information of Chinese characters. Specifically, for each character $x_i$ in the input sentence $X$, the encoder first produces its char embedding, pinyin embedding, and glyph embedding, all with embedding size $D$. These three embeddings are then concatenated and mapped to a $D$-dimensional fused embedding via a fully connected layer. After that, just like in most other pre-trained language models, the fused embedding is added with a position embedding, and fed into a stack of successive Transformer layers to generate a contextualized representation $\bm{h}_i \in \mathbb{R}^D$ for the input character $x_i$. We denote the character representations after this encoding process as $\bm{H}=$ $\{\bm{h}_1, \bm{h}_2, \cdots, \bm{h}_n\}$. As the encoder is not the main concern of this paper, we just provide a brief sketch of the encoder and refer readers to \citep{DBLP:conf/nips/VaswaniSPUJGKP17,DBLP:conf/acl/SunLSMAHWL20} for details.

\paragraph{Decoder for CSC}
This decoder is to predict the characters in the correct sentence $Y$ based on the encoding output $\bm{H}$. Specifically, given each input character $x_i$, we first project its encoding output $\bm{h}_i$ into a character-specific feature space:
\begin{equation}
\bm{h}^{(c)}_i ={\rm GeLU} \left( \bm{W}^{(c)} \bm{h}_i + \bm{b}^{(c)} \right),
\end{equation}
and then predict the corresponding correct character $\hat{y}_i$ based on the projection output:
\begin{equation}
p(\hat{y}_i|X) = {\rm softmax}\left( \bm{W}^{(y)} \bm{h}^{(c)}_i + \bm{b}^{(y)} \right).
\end{equation}
Here $\bm{W}^{(c)} \in \mathbb{R}^{D \times D}$, $\bm{b}^{(c)} \in \mathbb{R}^{D}$ are learnable parameters of the character-specific feature projection layer; $ \bm{W}^{(y)} \in \mathbb{R}^{V \times D}$, $\bm{b}^{(y)} \in \mathbb{R}^{V}$ are learnable parameters of the character prediction layer; $V$ is the vocabulary size. 

\paragraph{Decoder for CPP}
This decoder is to predict the fine-grained pinyin, {\it i.e.}, the initial, final, and tone, of each character in the correct sentence $Y$ based on the encoding output $\bm{H}$. Again, given each input character $x_i$ and its encoding output $\bm{h}_i$, we project $\bm{h}_i$ into a pronunciation-specific feature space:
\begin{equation}\label{eq:pron-projection}
\bm{h}^{(p)}_i ={\rm GeLU} \left( \bm{W}^{(p)} \bm{h}_i + \bm{b}^{(p)} \right),
\end{equation}
and predict the initial $\hat{\alpha}_i$, final $\hat{\beta}_i$, and tone $\hat{\gamma}_i$ of the corresponding correct character based on the projection output:
\begin{eqnarray}
p(\hat{\alpha}_i|X) &\!\!\!\!=\!\!\!\!& {\rm softmax}\left( \bm{W}^{(\alpha)} \bm{h}^{(p)}_i + \bm{b}^{(\alpha)} \right), \\
p(\hat{\beta}_i|X) &\!\!\!\!=\!\!\!\!& {\rm softmax}\left( \bm{W}^{(\beta)} \bm{h}^{(p)}_i + \bm{b}^{(\beta)} \right), \\
p(\hat{\gamma}_i|X) &\!\!\!\!=\!\!\!\!& {\rm softmax}\left( \bm{W}^{(\gamma)} \bm{h}^{(p)}_i + \bm{b}^{(\gamma)} \right).
\end{eqnarray}
Here $\bm{W}^{(p)} \in \mathbb{R}^{D \times D}$ and $\bm{b}^{(p)} \in \mathbb{R}^{D}$ are learnable parameters of the pronunciation-specific feature projection layer; $ \bm{W}^{(\delta)} \in \mathbb{R}^{U \times D}$, $\bm{b}^{(\delta)} \in \mathbb{R}^{U}$ with $\delta \in \{\alpha, \beta, \gamma\}$ are learnable parameters of the pronunciation prediction layers; $U$ is the total number of pronunciation units (initials, finals, and tones).

\paragraph{Adaptive Task Weighting}
We devise an adaptive task weighting scheme to balance the primary CSC and auxiliary CPP tasks during training. Given an input sentence $X$, the CSC task aims to match the predicted characters $\{\hat{y}_i\}_{i=1}^n$ with the ground truth $\{y_i\}_{i=1}^n$, while the CPP task aims to match the predicted fine-grained pinyin $\{(\hat{\alpha}_i, \hat{\beta}_i, \hat{\gamma}_i)\}_{i=1}^n$ with the ground truth $\{(\alpha_i, \beta_i, \gamma_i)\}_{i=1}^n$. Their loss functions are respectively defined as:
\begin{eqnarray}
\mathcal{L}^{(c)}_i &\!\!\!\!=\!\!\!\!& - \log p(\hat{y}_i=y_i|X), \label{eq:char-loss} \\
\mathcal{L}^{(p)}_i &\!\!\!\!=\!\!\!\!& - \frac{1}{3} \sum_{\delta \in \{\alpha, \beta, \gamma\}} \log p(\hat{\delta}_i=\delta_i|X), \label{eq:pron-loss}
\end{eqnarray}
where $\mathcal{L}^{(c)}_i$, $\mathcal{L}^{(p)}_i$ are the character and pronunciation prediction losses associated with the $i$-th character in the sentence, and the pronunciation prediction loss $\mathcal{L}^{(p)}_i$ is averaged over the initial, final, and tone prediction. 

Then as we have discussed earlier in the introduction, the auxiliary CPP task might provide different levels of benefits given different input characters. The more similar the input and target characters are in their pronunciation, the more likely there would be a spelling error caused by phonetic similarity. And to this case the CPP task might provide greater benefits and should be assigned a larger weight. To calculate such adaptive weights, we feed the target correct sentence $Y$ to the encoder and the followup pronunciation-specific projection layer. Then we calculate for each input character $x_i$ and its target character $y_i$ a cosine similarity ${\rm cos}(\bm{h}^{(p)}_{x_i}, \bm{h}^{(p)}_{y_i})$ based on their pronunciation-specific feature representations $\bm{h}^{(p)}_{x_i}$, $\bm{h}^{(p)}_{y_i}$ (see Eq.~(\ref{eq:pron-projection})), and accordingly define the adaptive weight at the $i$-th position as:
\begin{equation}\label{eq:weight}
w_i = e^{-({\rm cos}(\bm{h}^{(p)}_{x_i}, \bm{h}^{(p)}_{y_i})-1)^2}.
\end{equation}
The higher the cosine similarity ${\rm cos}(\bm{h}^{(p)}_{x_i}, \bm{h}^{(p)}_{y_i})$ is, the larger the weight $w_i$ will be. Finally, the overall loss is defined as the CSC loss with an adaptively weighted CPP loss: 
\begin{equation}\label{eq:loss}
\mathcal{L} =  \frac{1}{n} \sum_{i=1}^{n} \left( \mathcal{L}^{(c)}_i + w_i \mathcal{L}^{(p)}_i \right),
\end{equation}
where $\mathcal{L}^{(c)}_i$ and $\mathcal{L}^{(p)}_i$ are the character-specific CSC and CPP losses defined in Eq.~(\ref{eq:char-loss}) and Eq.~(\ref{eq:pron-loss}), respectively. There are two points worth noting here: (1) The branch of encoding and mapping the target sentence $Y$ is employed solely in the forward pass to calculate adaptive weights, and will be detached in the backward pass. (2) The auxiliary CPP task, as well as the adaptive weighting scheme, is introduced solely during training. At inference time, we use the CSC decoder alone for prediction.

\subsection{Constrained Iterative Correction}
\label{section:3.3}
As pointed out by \citet{liu-etal-2022-craspell}, advanced CSC models based on pre-trained language models ({\it e.g.}, BERT \citep{DBLP:conf/naacl/DevlinCLT19} and ChineseBERT \citep{DBLP:conf/acl/SunLSMAHWL20}) typically have poor performance on multi-typo texts, and tend to overcorrect valid expressions to more frequent expressions. To address these deficiencies, we devise a simple yet effective constrained iterative correction strategy during inference. Specifically, at inference time, for each input sentence we detect and correct spelling errors in an iterative fashion. During each iteration, only the corrections that appear in a specified window around each correction position in the previous iteration are allowed. After the iterations, if a position is modified every iteration, we restore this position to its original character without any correction. We empirically set the iteration number to 2 and the window size to 3 ({\it i.e.}, one position on the left and one on the right of the current position). As we will see later in our case study in Section~\ref{subsec:case}, this iterative correction strategy can effectively address the overcorrection issue and tackle difficult instances with multiple, in particular, consecutive errors.

\subsection{Further Pre-training with Confusion Set}
\label{section:3.4}
To obtain better initialization for SCOPE, we perform further pre-training by using a confusion set, as commonly practiced in most recently proposed CSC models \citep{DBLP:conf/acl/XuLZLWCHM21,DBLP:conf/acl/LiuYYZW20}. We consider wiki2019zh\footnote{\url{https://github.com/brightmart/nlp\_chinese\_corpus}} that consists of one million Chinese Wikipedia articles, split these articles into paragraphs, and regard each paragraph as a target sequence with no spelling errors. We further collect easily confused character pairs from a mixture of three publicly available confusion sets \citep{DBLP:conf/acl-sighan/WuLL13, lee2019building,DBLP:conf/emnlp/WangSLHZ18}, and retain only the pairs where both characters appear frequently (top 40\%) in the wiki2019zh corpus. Then, for each target sequence, we create a potentially misspelled input sequence by randomly selecting and replacing 15\% of the characters. Each selected character is replaced with an easily confused character (if any) 80\% of the time, a random character from the vocabulary 10\% of the time, and remained unchanged 10\% of the time. After that, we pre-train SCOPE on these misspelled and correct sequence pairs before adapting it to target datasets.

\section{Experiments and Results}
In this section, we introduce our experiments and results on SIGHAN benchmarks \citep{DBLP:conf/acl-sighan/WuLL13,DBLP:conf/acl-sighan/YuLTC14,DBLP:conf/acl-sighan/TsengLCC15}. We then verify the effectiveness of our model design, in particular the adaptivity and granularity of the auxiliary CPP task, via extensive ablation studies and analyses.

\subsection{Experimental Setups}\label{subsec:setups}

\begin{table}[t]
\small
\centering%\setlength{\tabcolsep}{5pt}
\begin{tabular*}{0.48 \textwidth}{@{\extracolsep{\fill}}lrrr}
\toprule
Training Set & \#Sent & Avg. Length & \#Errors \\
\midrule
SIGHAN15  & 2,338 & 31.1 & 3,037 \\
SIGHAN14  & 3,437 & 49.6 & 5,122 \\
SIGHAN13  & 700    & 41.8 & 343    \\
Wang271K  & 271,329 & 42.6 & 381,962 \\
%Total            & 277,804 & 42.6 & 390,464 \\
\bottomrule
\toprule
Test Set & \#Sent & Avg. Length & \#Errors \\
\midrule
SIGHAN15  & 1,100 & 30.6 & 703    \\
SIGHAN14  & 1,062 & 50.0 & 771    \\
SIGHAN13  & 1,000 & 74.3 & 1,224 \\
\bottomrule
\end{tabular*}
\caption{\label{data}Statistics of the datasets, including the number of sentences, the average length of sentences in tokens, and the number of errors in characters. We train on a combination of the training sets, and evaluate separately on each test set.}
\end{table}

\begin{table*}[t]
\small
\centering\setlength{\tabcolsep}{5pt}
\begin{tabular*}{\textwidth}{@{\extracolsep{\fill}}lccccccc}
\toprule
\multirow{2}{*}{Dataset}  & \multirow{2}{*}{Model}           & \multicolumn{3}{c}{Detection-level}          & \multicolumn{3}{c}{Correction-level}          \\ 
\cmidrule(r){3-5}\cmidrule(l){6-8}
                            &                                  & D-P           & D-R           & D-F           & C-P           & C-R           & C-F           \\ \midrule
\multirow{6}{*}{SIGHAN15}   & FASpell \citep{DBLP:conf/aclnut/HongYHLL19}             & 67.6          & 60.0          & 63.5          & 66.6          & 59.1          & 62.6          \\
                            & SpellGCN \citep{DBLP:conf/acl/ChengXCJWWCQ20}               & 74.8          & 80.7          & 77.7          & 72.1          & 77.7          & 75.9          \\
                            & MLM-phonetics \citep{DBLP:conf/acl/ZhangPZWHSWW21} & 77.5          & 83.1          & 80.2          & 74.9          & 80.2          & 77.5          \\
                            & REALISE \citep{DBLP:conf/acl/XuLZLWCHM21}       & 77.3          & 81.3          & 79.3          & 75.9          & 79.9          & 77.8          \\
                            & PLOME \citep{DBLP:conf/acl/LiuYYZW20}                  & 77.4          & 81.5          & 79.4          & 75.3          & 79.3          & 77.2          \\
                            & SCOPE (ours)                             & \textbf{81.1} & \textbf{84.3}              & \textbf{82.7} & \textbf{79.2} & \textbf{82.3} & \textbf{80.7} \\ 
\bottomrule
\toprule
\multirow{5}{*}{SIGHAN14} & FASpell \citep{DBLP:conf/aclnut/HongYHLL19}             & 61.0          & 53.5          & 57.0          & 59.4          & 52.0          & 55.4          \\
                        & SpellGCN \citep{DBLP:conf/acl/ChengXCJWWCQ20}               & 65.1          & 69.5          & 67.2          & 63.1          & 67.2          & 65.3          \\
                        & MLM-phonetics \citep{DBLP:conf/acl/ZhangPZWHSWW21} & 66.2          & \textbf{73.8} & 69.8          & 64.2          & \textbf{73.8} & 68.7          \\
                        & REALISE \citep{DBLP:conf/acl/XuLZLWCHM21}       & 67.8          & 71.5          & 69.6          & 66.3          & 70.0          & 68.1          \\
                        & SCOPE (ours)                             & \textbf{70.1} & 73.1          & \textbf{71.6}           & \textbf{68.6} & 71.5          & \textbf{70.1} \\ 
                        \bottomrule
                        \toprule
\multirow{5}{*}{SIGHAN13} & FASpell \citep{DBLP:conf/aclnut/HongYHLL19}             & 76.2          & 63.2          & 69.1          & 73.1          & 60.5          & 66.2          \\
                        & SpellGCN \citep{DBLP:conf/acl/ChengXCJWWCQ20}               & 80.1          & 74.4          & 77.2          & 78.3          & 72.7          & 75.4          \\
                        & MLM-phonetics \citep{DBLP:conf/acl/ZhangPZWHSWW21} & 82.0          & 78.3          & 80.1          & 79.5          & 77.0          & 78.2          \\
                        & REALISE \citep{DBLP:conf/acl/XuLZLWCHM21}$^\dag$       & \textbf{88.6} & 82.5          & \textbf{85.4} & \textbf{87.2} & 81.2          & 84.1          \\
                        & SCOPE (ours)$^\dag$                             & 87.4          & \textbf{83.4} & \textbf{85.4} & 86.3          & \textbf{82.4} & \textbf{84.3} \\ 
\bottomrule   
\end{tabular*}
\caption{\label{main result}Sentence-level performance on the test sets of SIGHAN13, SIGHAN14, SIGHAN15, where precision (P), recall (R), F1 (F) for detection (D) and correction (C) are reported ($\%$). Baseline results are directly taken from their respective literatures. Results marked by ``$\dag$'' are obtained by applying a post-processing step on SIGHAN13 which removes all detected and corrected ``的'', ``地'', ``得'' from the model output before evaluation, due to the relatively poor annotation quality about ``的'', ``地'', ``得'' on SIGHAN13 as observed and suggested by \citet{DBLP:conf/acl/XuLZLWCHM21}.}
\end{table*}

\paragraph{Datasets and Evaluation Metrics}
As in previous work \citep{DBLP:conf/acl/ChengXCJWWCQ20,DBLP:conf/acl/LiuYYZW20,DBLP:conf/acl/XuLZLWCHM21}, our training data is a combination of (1) manually annotated training examples from SIGHAN13 \citep{DBLP:conf/acl-sighan/WuLL13}, SIGHAN14 \citep{DBLP:conf/acl-sighan/YuLTC14}, SIGHAN15 \cite{DBLP:conf/acl-sighan/TsengLCC15}, and (2) 271K training examples from \citet{DBLP:conf/emnlp/WangSLHZ18} automatically generated by ASR- and OCR-based methods. We employ the test sets of SIGHAN13, SIGHAN14, SIGHAN15 for evaluation. The statistics of the used datasets are shown in Table \ref{data}. The original SIGHAN datasets are in traditional Chinese. We follow previous work \citep{DBLP:conf/acl/ChengXCJWWCQ20,DBLP:conf/acl/XuLZLWCHM21} to convert them to simplified Chinese using OpenCC\footnote{\url{https://github.com/BYVoid/OpenCC}}. We further use pypinyin\footnote{\url{https://pypi.org/project/pypinyin}} to obtain the pinyin of each character, and segment it into the initial, final, and tone using a pre-defined vocabulary of initials and finals provided by \citet{DBLP:conf/acl/XuLZLWCHM21}.\footnote{\url{https://github.com/DaDaMrX/ReaLiSe}}

We use the widely adopted sentence-level precision, recall and F1 as our main evaluation metrics. Sentence-level metrics are stricter than character-level metrics since a sentence is considered to be correct if and only if all errors in the sentence are successfully detected and corrected. Metrics are reported on the detection and correction sub-tasks. Besides sentence-level evaluation, we also consider character-level evaluation and the official SIGHAN evaluation.  We leave their results to Appendix~\ref{sec:other evaluation}

\paragraph{Baseline Methods}
We compare SCOPE against the following baseline methods. All these methods have employed character phonetic information in some manner, and represent current state-of-the-art on the SIGHAN benchmarks.
\begin{itemize}
    \item {\it FASpell} \citep{DBLP:conf/aclnut/HongYHLL19} employs BERT to generate candidates for corrections and filters visually/phonologically irrelevant candidates by a confidence-similarity decoder.
    \item {\it SpellGCN} \citep{DBLP:conf/acl/ChengXCJWWCQ20} learns pronunciation/shape similarities between characters via GCNs, and combines the graph representations with BERT output for final prediction.
    \item {\it MLM-phonetics} \citep{DBLP:conf/acl/ZhangPZWHSWW21} jointly fine-tunes a detection module and a correction module on the basis of a pre-trained language model with phonetic features.
    \item {\it REALISE} \citep{DBLP:conf/acl/XuLZLWCHM21} models semantic, phonetic and visual information of input characters, and selectively mixes information in these modalities to predict final corrections.
    \item {\it PLOME} \citep{DBLP:conf/acl/LiuYYZW20} extracts phonetic and visual features of characters using GRU. It also predicts the pronunciation of target characters, but in a coarse-grained, non-adaptive manner.
\end{itemize}

\paragraph{Implementation Details}
In SCOPE, the encoder is initialized from ChineseBERT-base\footnote{\url{https://github.com/ShannonAI/ChineseBert}}, while the decoders are randomly initialized. We then conduct further pre-training on wiki2019zh for 1 epoch with a batch size of $512$ and a learning rate of $10^{-4}$. The other hyperparameters are set to their default values as in ChineseBERT \citep{DBLP:conf/acl/SunLSMAHWL20}. During this pre-training stage, we do not use the adaptive task weighting scheme, and simply set the auxiliary CPP task weight to 1 for all characters for computational efficiency. After that, we fine-tune on the combined training set. We set the maximum sequence length to $192$ and the learning rate to $5\!\times\! 10^{-5}$. The optimal models on SIGHAN13/SIGHAN14/SIGHAN15 are obtained by training with batch sizes of $96/96/$ $64$ for $20/30/30$ epochs, respectively. Other hyperparameters are again set to their default values as in ChineseBERT. All experiments are conducted on 2 GeForce RTX 3090 with 24G memory. 

\subsection{Main Results}
Table \ref{main result} presents the sentence-level performance of SCOPE and its baseline methods on the test sets of SIGHAN13, SIGHAN14, and SIGHAN15. We can see that SCOPE consistently outperforms all the baselines on all the datasets in almost all metrics, verifying its effectiveness and superiority for CSC. The improvements, in most cases, are rather substantial, {\it e.g.}, +2.5/+2.9 detection/correction F1 on SIGHAN15 and +1.8/+1.4 detection/correction F1 on SIGHAN14. Note that on SIGHAN13, although the improvements over the best performing baseline REALISE are somehow limited, SCOPE still outperforms the second best performing baseline MLM-phonetics by large margins (+5.3/+6.1 detection/correction F1). We attribute this phenomenon to the fact that the annotation quality is relatively poor on SIGHAN13, with a lot of mixed usage of ``的'', ``地'', ``得'' not annotated \citep{DBLP:conf/acl/ChengXCJWWCQ20}. We hence follow REALISE \citep{DBLP:conf/acl/XuLZLWCHM21} and remove all detected and corrected ``的'', ``地'', ``得'' from the model output before evaluation. This post-processing trick is extremely useful on SIGHAN13, and it might even conceal improvements from other strategies on this dataset. 

Besides sentence-level metrics, we also consider character-level evaluation and the official SIGHAN evaluation, and make further comparison to some other methods that have their results reported in these settings \citep{DBLP:conf/emnlp/JiYQ21,liu-etal-2022-craspell}. We leave the results to Appendix~\ref{sec:other evaluation}, which reveal that SCOPE still performs the best in these new settings.

\subsection{Eliminating Encoder Differences}
As SCOPE uses a different and potentially more powerful encoder ({\it i.e.}, ChineseBERT) compared to the baselines, we further conduct experiments to eliminate the effects of different encoders and focus solely on the auxiliary CPP task, which is the main contribution of this work. To do so, we initialize the encoder from a well-trained REALISE model (one of the best performing baselines with its code and model released to the public). Then, we perform further pre-training on wiki2019zh and fine-tune on the combination of SIGHAN benchmarks with our adaptively-weighted, fine-grained CPP task. The pre-training and fine-tuning configurations are the same as those introduced above in Section~\ref{subsec:setups}. The constrained iterative correction (CIC) strategy is also applied during inference. We call this setting SCOPE (REALISE).

\begin{table}[t]
\small
\centering\setlength{\tabcolsep}{4pt}
\begin{tabular*}{0.48 \textwidth}{@{\extracolsep{\fill}}lcccccc}
\toprule
& \multicolumn{3}{c}{Detection-level} & \multicolumn{3}{c}{Correction-level} \\  
\cmidrule(lr){2-4}\cmidrule(lr){5-7}
& D-P & D-R & D-F & C-P & C-R & C-F \\
\midrule
SIGHAN15 \\
REALISE & 77.3 & 81.3 & 79.3 & 75.9 & 79.9 & 77.8 \\
SCOPE (REALISE) & \textbf{78.7} & \textbf{84.7} & \textbf{81.6} & \textbf{76.8} & \textbf{82.6} & \textbf{79.6} \\
\bottomrule
\toprule
SIGHAN14 \\
REALISE & 67.8 & 71.5 & 69.6 & 66.3 & 70.0 & 68.1 \\
SCOPE (REALISE) & \textbf{69.0} & \textbf{75.0} & \textbf{71.9} & \textbf{67.1} & \textbf{72.9} & \textbf{69.9} \\
\bottomrule
\toprule
SIGHAN13 \\
REALISE & \textbf{88.6} & 82.5 & \textbf{85.4} & \textbf{87.2} & 81.2 & 84.1 \\
SCOPE (REALISE) & 87.5 & \textbf{83.2} & 85.3 & 86.4 & \textbf{82.3} & \textbf{84.3} \\
\bottomrule
\end{tabular*}
\caption{\label{tab: Encoder}Performance of SCOPE with the same encoder as REALISE on test sets of SIGHAN13, SIGHAN14, and SIGHAN15.}
\end{table}

Table~\ref{tab: Encoder} presents the sentence-level performance of this new setting on the test sets of SIGHAN13, SIGHAN14, and SIGHAN15. We can observe that SCOPE (REALISE) consistently outperforms its direct opponent REALISE on all the datasets. The improvements, in most cases, are rather substantial, except for those on the relatively poorly annotated SIGHAN13. These results verify the effectiveness of our approach irrespective of the encoder.

\subsection{Effects of Adaptivity and Granularity}
This section then investigates the effects of {\it adaptivity} and {\it granularity} of the auxiliary CPP task on the overall CSC performance. 

\paragraph{Adaptivity}
As for adaptivity, we make comparison among the following three diverse task weighting schemes that balance the CSC and CPP tasks.
\begin{itemize}
\item {\it Fully-adaptive} (Full-adapt) is the scheme we used in SCOPE. It determines the CPP task weights according to phonological similarities between input and target characters, and the similarities are further adjusted dynamically during model training (see Eq.~(\ref{eq:weight})).
\item {\it Partially-adaptive} (Part-adapt) also decides the CPP task weights according to phonological similarities, but the similarities are static, defined as $w_i=1 - {\rm norm}({\rm edit\_distance}_i)$, where ${\rm edit\_distance}_i$ is the Levenshtein edit distance \citep{levenshtein1966binary} between the pinyin sequences of the $i$ input and target characters and ${\rm norm}(\cdot)$ is a normalization function. The smaller the edit distance is, the larger the weight will be.
\item {\it Non-adaptive} (Non-adapt) considers no adaptivity and simply sets the CPP task weight to 1 for all characters ($w_i=1$ for all $i$).
\end{itemize}
We compare the three settings in the SIGHAN fine-tuning stage, starting from the same checkpoint after pre-training on wiki2019zh with a non-adaptive task weighting scheme. Here Full-adapt is equivalent to SCOPE. 

\begin{table}[t]
\small
\centering\setlength{\tabcolsep}{4pt}
\begin{tabular*}{0.48 \textwidth}{@{\extracolsep{\fill}}lcccccc}
\toprule
\multirow{2}{*}{SIGHAN15} & \multicolumn{3}{c}{Detection-level} & \multicolumn{3}{c}{Correction-level} \\  
\cmidrule(lr){2-4}\cmidrule(lr){5-7}
& D-P & D-R & D-F & C-P & C-R & C-F \\
\midrule
REALISE & 77.3 & 81.3 & 79.3 & 75.9 & 79.9 & 77.8 \\
%SCOPE   & 81.1 & 84.3 & 82.7 & 79.2 & 82.3 & 80.7 \\
w/o CPP  & 79.1 & 82.4 & 80.7 & 76.8 & 80.0 & 78.4 \\
\bottomrule
\toprule
\multicolumn{7}{c}{Effects of Adaptivity} \\
\midrule
Non-adapt & 79.0 & 83.5 & 81.2 & 76.6 & 81.0 & 78.7 \\
Part-adapt & 80.1 & 83.5 & 81.8 & 78.0 & 81.3 & 79.6 \\
Full-adapt  & 81.1 & 84.3 & 82.7 & 79.2 & 82.3 & 80.7 \\
%Full-adapt  & {\bf 81.1} & {\bf 84.3} & {\bf 82.7} & {\bf 79.2} & {\bf 82.3} & {\bf 80.7} \\
\bottomrule
\toprule
\multicolumn{7}{c}{Effects of Granularity} \\
\midrule
Coarse      & 79.9 & 83.7 & 81.8 & 77.4 & 81.1 & 79.2 \\
Fine           & 81.1 & 84.3 & 82.7 & 79.2 & 82.3 & 80.7 \\
%Fine-grain           & {\bf 81.1} & {\bf 84.3} & {\bf 82.7} & {\bf 79.2} & {\bf 82.3} & {\bf 80.7} \\ 
\bottomrule
\end{tabular*}
\caption{\label{tab: Adaptivity}Performance of SCOPE with different levels of adaptivity and granularity of the auxiliary CPP task on the test set of SIGHAN15.}% REALISE and SCOPE w/o CPP results are also provided for reference.
\end{table}

\paragraph{Granularity}
As for granularity, we consider and make comparison between two types of CPP tasks.
\begin{itemize}
\item {\it Fine-grained} (Fine) is the task we employed in SCOPE that predicts the initial, final, and tone of the pinyin of each target character.
\item {\it Coarse-grained} (Coarse) is a task that predicts the whole pinyin of each target character.
\end{itemize}
For fair comparison, we also introduce further pre-training on wiki2019zh with a coarse-grained CPP task, and use this checkpoint to initialize the Coarse setting during SIGHAN fine-tuning. In both two settings the CPP task is adaptively weighted as in Eq.~(\ref{eq:weight}), and Fine is equivalent to SCOPE.

\paragraph{Results}
Table \ref{tab: Adaptivity} presents the sentence-level performance of these SCOPE variants on the test set of SIGHAN15. The scores of our best performing baseline REALISE as well as SCOPE without the CPP task (denoted as w/o CPP) are also provided for reference. We can see that introducing an auxiliary CPP task always brings benefits to CSC, no matter what level of adaptivity and granularity the task is. As for the adaptivity of task weighting, the Full-adapt scheme that considers dynamic adaptivity performs better than Part-adapt that considers static adaptivity, which in turn performs better than Non-adapt that considers no adaptivity. As for the granularity, a fine-grained CPP task performs better than a coarse-grained one. These results verify the rationality of introducing a fine-grained CPP task with adaptive task weighting to improve CSC.

\subsection{Ablation and Case Study}\label{subsec:case}

\paragraph{Ablation Study}
We conduct ablation studies on SIGHAN15 with the following settings: (1) removing the auxiliary CPP task (w/o CPP); (2) removing further pre-training on wiki2019zh (w/o FPT); and (3) removing the constrained iterative correction strategy at inference time (w/o CIC). The results are presented in Table~\ref{tab:ablation}. We can see that no matter which component we remove, the performance of SCOPE drops. This fully demonstrates the effectiveness of each component in our method.

\begin{table}[t]
\small
\centering\setlength{\tabcolsep}{4pt}
\begin{tabular*}{0.48 \textwidth}{@{\extracolsep{\fill}}lcccccc}
\toprule
\multirow{2}{*}{SIGHAN15} & \multicolumn{3}{c}{Detection-level} & \multicolumn{3}{c}{Correction-level} \\  
\cmidrule(lr){2-4}\cmidrule(lr){5-7}
& D-P & D-R & D-F & C-P & C-R & C-F \\
\midrule
SCOPE   & 81.1 & 84.3 & 82.7 & 79.2 & 82.3 & 80.7 \\
w/o CPP  & 79.1 & 82.4 & 80.7 & 76.8 & 80.0 & 78.4 \\
w/o FPT & 80.2 & 83.2 & 81.7 & 77.5 & 80.4 & 78.9\\
w/o CIC & 78.3 & 82.6 & 80.4 & 76.5 & 80.8 & 78.6\\
\bottomrule
\end{tabular*}
\caption{\label{tab:ablation}Ablation results on the test set of SIGHAN15. The following changes are applied to SCOPE: removing the CPP task (w/o CPP), removing further pre-training (w/o FPT), and removing constrained iterative correction (w/o CIC).}
\end{table}

\begin{table}[t]
\small
\centering\setlength{\tabcolsep}{5pt}
\begin{tabular*}{0.48 \textwidth}{@{\extracolsep{\fill}}ll}
\toprule
\multicolumn{2}{@{}l}{Tackle Consecutive Errors}  \\
\toprule
Input: & $\!\!\!\!$	我以前想要\textcolor{red}{高}诉你，可我忘了。我真\textcolor{red}{户秃}。   \\
                        & \makecell[l]{$\!\!\!\!$	I tried to high you before, but I forgot. I'm really \\$\!\!\!\!$ house bald.}   \\ 
Iteration 1:  & $\!\!\!\!$   我以前想要\textcolor{blue}{\underline{告}}诉你，可我忘了。我真\textcolor{red}{户}\textcolor{blue}{\underline{涂}}。 \\
              & \makecell[l]{$\!\!\!\!$   I tried to tell you before, but I forgot. I'm really \\$\!\!\!\!$ house painted.} \\
Iteration 2:  & $\!\!\!\!$   我以前想要告诉你，可我忘了。我真\textcolor{blue}{\underline{糊}}涂。  \\
              & \makecell[l]{$\!\!\!\!$   I tried to tell you before, but I forgot. I'm really \\$\!\!\!\!$ muddled.} \\
\midrule
Input: & $\!\!\!\!$	可是\textcolor{red}{福物}生对我们很客气。   \\
      & $\!\!\!\!$	But the fortune object man was polite to us.  \\
Iteration 1:  & $\!\!\!\!$   可是\textcolor{red}{福}\textcolor{blue}{\underline{务}}生对我们很客气。 \\
              & $\!\!\!\!$   But the fortune business man was polite to us. \\
Iteration 2:  & $\!\!\!\!$   可是\textcolor{blue}{\underline{服}}务生对我们很客气。  \\
              & $\!\!\!\!$   But the waiter was very polite to us.  \\
\bottomrule
\toprule
\multicolumn{2}{@{}l}{Address Over-correction Issue}  \\
\toprule
Input: & $\!\!\!\!$  他再也不会\textcolor{red}{撤扬}。   \\
        & $\!\!\!\!$  He will never withdraw raise again.   \\
Iteration 1:   & $\!\!\!\!$   \textcolor{red}{\underline{她}}再也不会\textcolor{red}{撤}\textcolor{blue}{\underline{样}}。  \\
              & $\!\!\!\!$   She will never withdraw appearance again.  \\
Iteration 2:  & $\!\!\!\!$   \textcolor{blue}{\underline{他}}再也不会\textcolor{blue}{\underline{这}}样。  \\
              & $\!\!\!\!$   He will never do this again.  \\
\midrule
Input: & $\!\!\!\!$	幸运地，隔天她带着辞典来学校。   \\
        & \makecell[l]{$\!\!\!\!$   Fortunately, she came to school the next day \\$\!\!\!\!$ with a thesaurus.}   \\
Iteration 1:  & $\!\!\!\!$   幸运地，\textcolor{red}{\underline{葛}}天她带着辞典来学校。 \\
              & \makecell[l]{$\!\!\!\!$   Fortunately, Ge Tian she came to school with \\$\!\!\!\!$ a thesaurus.} \\
Iteration 2:  & $\!\!\!\!$   幸运地，\textcolor{blue}{\underline{隔}}天她带着辞典来学校。  \\
              & \makecell[l]{$\!\!\!\!$   Fortunately, she came to school the next day \\$\!\!\!\!$ with a thesaurus.}  \\
\bottomrule
\end{tabular*}
\caption{\label{case study}Cases from the SIGHAN15 test set to show how the iterative correction strategy can tackle consecutive errors and address the over-correction issue. Erroneous characters are in red, and their SCOPE corrections are in blue and underlined.} 
\end{table}

\paragraph{Case Study}
Table~\ref{case study} further shows several cases from the SIGHAN15 test set to illustrate how the constrained iterative correction strategy (see Section~\ref{section:3.3}) can effectively tackle consecutive spelling errors and address the over-correction issue. For consecutive errors, {\it e.g.}, ``户秃'' in the first case, this strategy is able to correct them iteratively, one character at a time, {\it e.g.}, by modifying ``秃'' to ``涂'' in the first round and then ``户'' to ``糊'' in the second round. For over-correction where the model makes unnecessary modifications, {\it e.g.}, ``他'' to ``她'' in the third case and ``隔'' to ``葛'' in the fourth case, the iterative correction strategy can always change them back most of the time.

\section{Conclusions}
This paper proposes SCOPE, which employs a fine-grained Chinese pronunciation prediction (CPP) task with adaptive task weighting to improve the performance of Chinese spelling check (CSC). Our method builds upon a shared encoder two parallel decoders, one to predict target characters {\it i.e.}, CSC, and the other to predict initials, finals, and tones of target characters, {\it i.e.}, fine-grained CPP. The two decoders are then balanced adaptively according to the phonetic similarity between input and target characters. An iterative correction strategy is further designed during inference. SCOPE establishes new state-of-the-art on three SIGHAN benchmarks, verifying the effectiveness and superiority of introducing an auxiliary CPP task to improve CSC. Extensive ablation studies further verify the positive effects of dynamic adaptivity and fine granularity of this auxiliary task. %To our knowledge, this is the first work that systematically discusses the adaptivity and granularity of the auxiliary CPP task. 

%\begin{figure}[t]
%\centering
%\includegraphics[scale=0.45]{pplm.pdf}
%\caption{Performance improvement brought by CPP task. Here, step size is a multiplicative factor of gradient on universal feature, 
%and D-F, C-F correspond to F1 scores of detection and correction level. }
%\label{fig:2}
%\end{figure}

\section*{Limitations}
SCOPE introduces an auxiliary CPP task alongside the primary CSC task in the training phase. This auxiliary CPP task causes 28\% extra overhead of computation, with the runtime per epoch increasing from $19.32$ minutes to $24.68$ minutes. But the extra overhead of GPU memory is almost negligible, as the CPP decoder contains only $1$M out of the total $148$M parameters of the whole model (to which the encoder contributes $146$M parameters). Note that the additional overhead caused by CPP is required only in the training phase, but not at inference time.

\section*{Acknowledgements}
We would like to thank all the reviewers for their insightful and valuable suggestions, which significantly improve the quality of this paper. This work is supported by National Natural Science Fundation of China under Grants 61876223, 62222212 and U19A2057, and Science Fund for Creative Research Groups under Grant 62121002. 

%Besides, the proposed CPP-based method only experiments on ChineseBERT attached with phonetic and glyph encoding, which is essential for CSC. 
%And compared to those advanced models \citep{DBLP:conf/acl/XuLZLWCHM21, DBLP:conf/acl/HuangLJZCWX20} for CSC, which require additional pre-training of the phonetic and glyph encoders based on the pre-trained model, ChineseBERT pre-trains all encoders together to be convenient for use. 
%It is planned to apply our method to those advanced models in future work to verify its effectiveness.  

\bibliography{custom}

\begin{thebibliography}{26}
\expandafter\ifx\csname natexlab\endcsname\relax\def\natexlab#1{#1}\fi

\bibitem[{Chen et~al.(2013)Chen, Lee, Lee, Wang, and
  Chen}]{DBLP:conf/acl-sighan/ChenLLWC13}
Kuan{-}Yu Chen, Hung{-}Shin Lee, Chung{-}Han Lee, Hsin{-}Min Wang, and
  Hsin{-}Hsi Chen. 2013.
\newblock \href {https://aclanthology.org/W13-4414/} {A study of language
  modeling for chinese spelling check}.
\newblock In \emph{Proceedings of the Seventh {SIGHAN} Workshop on Chinese
  Language Processing, SIGHAN@IJCNLP 2013, Nagoya, Japan, October 14-18, 2013},
  pages 79--83. Asian Federation of Natural Language Processing.

\bibitem[{Cheng et~al.(2020)Cheng, Xu, Chen, Jiang, Wang, Wang, Chu, and
  Qi}]{DBLP:conf/acl/ChengXCJWWCQ20}
Xingyi Cheng, Weidi Xu, Kunlong Chen, Shaohua Jiang, Feng Wang, Taifeng Wang,
  Wei Chu, and Yuan Qi. 2020.
\newblock \href {https://doi.org/10.18653/v1/2020.acl-main.81} {Spellgcn:
  Incorporating phonological and visual similarities into language models for
  chinese spelling check}.
\newblock In \emph{Proceedings of the 58th Annual Meeting of the Association
  for Computational Linguistics, {ACL} 2020, Online, July 5-10, 2020}, pages
  871--881. Association for Computational Linguistics.

\bibitem[{Devlin et~al.(2019)Devlin, Chang, Lee, and
  Toutanova}]{DBLP:conf/naacl/DevlinCLT19}
Jacob Devlin, Ming{-}Wei Chang, Kenton Lee, and Kristina Toutanova. 2019.
\newblock \href {https://doi.org/10.18653/v1/n19-1423} {{BERT:} pre-training of
  deep bidirectional transformers for language understanding}.
\newblock In \emph{Proceedings of the 2019 Conference of the North American
  Chapter of the Association for Computational Linguistics: Human Language
  Technologies, {NAACL-HLT} 2019, Minneapolis, MN, USA, June 2-7, 2019, Volume
  1 (Long and Short Papers)}, pages 4171--4186. Association for Computational
  Linguistics.

\bibitem[{Guo et~al.(2021)Guo, Ni, Wang, Zhu, and
  Xie}]{DBLP:conf/acl/GuoNWZX21}
Zhao Guo, Yuan Ni, Keqiang Wang, Wei Zhu, and Guotong Xie. 2021.
\newblock \href {https://doi.org/10.18653/v1/2021.findings-acl.122} {Global
  attention decoder for chinese spelling error correction}.
\newblock In \emph{Findings of the Association for Computational Linguistics:
  {ACL/IJCNLP} 2021, Online Event, August 1-6, 2021}, volume {ACL/IJCNLP} 2021
  of \emph{Findings of {ACL}}, pages 1419--1428. Association for Computational
  Linguistics.

\bibitem[{Hong et~al.(2019)Hong, Yu, He, Liu, and
  Liu}]{DBLP:conf/aclnut/HongYHLL19}
Yuzhong Hong, Xianguo Yu, Neng He, Nan Liu, and Junhui Liu. 2019.
\newblock \href {https://doi.org/10.18653/v1/D19-5522} {Faspell: {A} fast,
  adaptable, simple, powerful chinese spell checker based on dae-decoder
  paradigm}.
\newblock In \emph{Proceedings of the 5th Workshop on Noisy User-generated
  Text, W-NUT@EMNLP 2019, Hong Kong, China, November 4, 2019}, pages 160--169.
  Association for Computational Linguistics.

\bibitem[{Huang et~al.(2021)Huang, Li, Jiang, Zhang, Chen, Wang, and
  Xiao}]{DBLP:conf/acl/HuangLJZCWX20}
Li~Huang, Junjie Li, Weiwei Jiang, Zhiyu Zhang, Minchuan Chen, Shaojun Wang,
  and Jing Xiao. 2021.
\newblock \href {https://doi.org/10.18653/v1/2021.acl-long.464} {Phmospell:
  Phonological and morphological knowledge guided chinese spelling check}.
\newblock In \emph{Proceedings of the 59th Annual Meeting of the Association
  for Computational Linguistics and the 11th International Joint Conference on
  Natural Language Processing, {ACL/IJCNLP} 2021, (Volume 1: Long Papers),
  Virtual Event, August 1-6, 2021}, pages 5958--5967. Association for
  Computational Linguistics.

\bibitem[{Ji et~al.(2021)Ji, Yan, and Qiu}]{DBLP:conf/emnlp/JiYQ21}
Tuo Ji, Hang Yan, and Xipeng Qiu. 2021.
\newblock \href {https://doi.org/10.18653/v1/2021.emnlp-main.287} {Spellbert:
  {A} lightweight pretrained model for chinese spelling check}.
\newblock In \emph{Proceedings of the 2021 Conference on Empirical Methods in
  Natural Language Processing, {EMNLP} 2021, Virtual Event / Punta Cana,
  Dominican Republic, 7-11 November, 2021}, pages 3544--3551. Association for
  Computational Linguistics.

\bibitem[{Jiang et~al.(2012)Jiang, Wang, Lin, Wang, Cheng, Liu, Wang, and
  Zhang}]{jiang2012rule}
Ying Jiang, Tong Wang, Tao Lin, Fangjie Wang, Wenting Cheng, Xiaofei Liu,
  Chenghui Wang, and Weijian Zhang. 2012.
\newblock \href {https://ieeexplore.ieee.org/abstract/document/6257223} {A rule
  based chinese spelling and grammar detection system utility}.
\newblock In \emph{2012 International Conference on System Science and
  Engineering (ICSSE)}, pages 437--440. IEEE.

\bibitem[{Lee et~al.(2019)Lee, Wu, Li, Lin, and Tseng}]{lee2019building}
Lung~Hao Lee, Wun~Syuan Wu, Jian~Hong Li, Yu~Chi Lin, and Yuen~Hsien Tseng.
  2019.
\newblock \href
  {https://www.researchgate.net/profile/Lung-Hao-Lee-2/publication/339068842_Building_a_Confused_Character_Set_for_Chinese_Spell_Checking/links/5e3bd25e458515072d831da6/Building-a-Confused-Character-Set-for-Chinese-Spell-Checking.pdf}
  {Building a confused character set for chinese spell checking}.
\newblock In \emph{27th International Conference on Computers in Education,
  ICCE 2019}, pages 703--705. Asia-Pacific Society for Computers in Education.

\bibitem[{Levenshtein et~al.(1966)}]{levenshtein1966binary}
Vladimir~I Levenshtein et~al. 1966.
\newblock \href {https://nymity.ch/sybilhunting/pdf/Levenshtein1966a.pdf}
  {Binary codes capable of correcting deletions, insertions, and reversals}.
\newblock In \emph{Soviet physics doklady}, volume~10, pages 707--710. Soviet
  Union.

\bibitem[{Liu et~al.(2011)Liu, Lai, Tien, Chuang, Wu, and
  Lee}]{DBLP:journals/talip/LiuLTCWL11}
C.{-}L. Liu, M.{-}H. Lai, Kan{-}Wen Tien, Y.{-}H. Chuang, Shih{-}Hung Wu, and
  C.{-}Y. Lee. 2011.
\newblock \href {https://doi.org/10.1145/1967293.1967297} {Visually and
  phonologically similar characters in incorrect chinese words: Analyses,
  identification, and applications}.
\newblock \emph{{ACM} Trans. Asian Lang. Inf. Process.}, 10(2):10:1--10:39.

\bibitem[{Liu et~al.(2022)Liu, Song, Yue, Yang, Cai, Yu, and
  Sun}]{liu-etal-2022-craspell}
Shulin Liu, Shengkang Song, Tianchi Yue, Tao Yang, Huihui Cai, TingHao Yu, and
  Shengli Sun. 2022.
\newblock \href {https://doi.org/10.18653/v1/2022.findings-acl.237}
  {{CRAS}pell: A contextual typo robust approach to improve {C}hinese spelling
  correction}.
\newblock In \emph{Findings of the Association for Computational Linguistics:
  ACL 2022}, pages 3008--3018, Dublin, Ireland. Association for Computational
  Linguistics.

\bibitem[{Liu et~al.(2021)Liu, Yang, Yue, Zhang, and
  Wang}]{DBLP:conf/acl/LiuYYZW20}
Shulin Liu, Tao Yang, Tianchi Yue, Feng Zhang, and Di~Wang. 2021.
\newblock \href {https://doi.org/10.18653/v1/2021.acl-long.233} {{PLOME:}
  pre-training with misspelled knowledge for chinese spelling correction}.
\newblock In \emph{Proceedings of the 59th Annual Meeting of the Association
  for Computational Linguistics and the 11th International Joint Conference on
  Natural Language Processing, {ACL/IJCNLP} 2021, (Volume 1: Long Papers),
  Virtual Event, August 1-6, 2021}, pages 2991--3000. Association for
  Computational Linguistics.

\bibitem[{Liu et~al.(2013)Liu, Cheng, Luo, Duh, and
  Matsumoto}]{DBLP:conf/acl-sighan/LiuCLDM13}
Xiaodong Liu, Kevin Cheng, Yanyan Luo, Kevin Duh, and Yuji Matsumoto. 2013.
\newblock \href {https://aclanthology.org/W13-4409/} {A hybrid chinese spelling
  correction using language model and statistical machine translation with
  reranking}.
\newblock In \emph{Proceedings of the Seventh {SIGHAN} Workshop on Chinese
  Language Processing, SIGHAN@IJCNLP 2013, Nagoya, Japan, October 14-18, 2013},
  pages 54--58. Asian Federation of Natural Language Processing.

\bibitem[{Mangu and Brill(1997)}]{DBLP:conf/icml/ManguB97}
Lidia Mangu and Eric Brill. 1997.
\newblock Automatic rule acquisition for spelling correction.
\newblock In \emph{Proceedings of the Fourteenth International Conference on
  Machine Learning {(ICML} 1997), Nashville, Tennessee, USA, July 8-12, 1997},
  pages 187--194. Morgan Kaufmann.

\bibitem[{Sun et~al.(2021)Sun, Li, Sun, Meng, Ao, He, Wu, and
  Li}]{DBLP:conf/acl/SunLSMAHWL20}
Zijun Sun, Xiaoya Li, Xiaofei Sun, Yuxian Meng, Xiang Ao, Qing He, Fei Wu, and
  Jiwei Li. 2021.
\newblock \href {https://doi.org/10.18653/v1/2021.acl-long.161} {Chinesebert:
  Chinese pretraining enhanced by glyph and pinyin information}.
\newblock In \emph{Proceedings of the 59th Annual Meeting of the Association
  for Computational Linguistics and the 11th International Joint Conference on
  Natural Language Processing, {ACL/IJCNLP} 2021, (Volume 1: Long Papers),
  Virtual Event, August 1-6, 2021}, pages 2065--2075. Association for
  Computational Linguistics.

\bibitem[{Tseng et~al.(2015)Tseng, Lee, Chang, and
  Chen}]{DBLP:conf/acl-sighan/TsengLCC15}
Yuen{-}Hsien Tseng, Lung{-}Hao Lee, Li{-}Ping Chang, and Hsin{-}Hsi Chen. 2015.
\newblock \href {https://doi.org/10.18653/v1/W15-3106} {Introduction to
  {SIGHAN} 2015 bake-off for chinese spelling check}.
\newblock In \emph{Proceedings of the Eighth {SIGHAN} Workshop on Chinese
  Language Processing, SIGHAN@IJCNLP 2015, Beijing, China, July 30-31, 2015},
  pages 32--37. Association for Computational Linguistics.

\bibitem[{Vaswani et~al.(2017)Vaswani, Shazeer, Parmar, Uszkoreit, Jones,
  Gomez, Kaiser, and Polosukhin}]{DBLP:conf/nips/VaswaniSPUJGKP17}
Ashish Vaswani, Noam Shazeer, Niki Parmar, Jakob Uszkoreit, Llion Jones,
  Aidan~N. Gomez, Lukasz Kaiser, and Illia Polosukhin. 2017.
\newblock \href
  {https://proceedings.neurips.cc/paper/2017/hash/3f5ee243547dee91fbd053c1c4a845aa-Abstract.html}
  {Attention is all you need}.
\newblock In \emph{Advances in Neural Information Processing Systems 30: Annual
  Conference on Neural Information Processing Systems 2017, December 4-9, 2017,
  Long Beach, CA, {USA}}, pages 5998--6008.

\bibitem[{Wang et~al.(2018)Wang, Song, Li, Han, and
  Zhang}]{DBLP:conf/emnlp/WangSLHZ18}
Dingmin Wang, Yan Song, Jing Li, Jialong Han, and Haisong Zhang. 2018.
\newblock \href {https://doi.org/10.18653/v1/d18-1273} {A hybrid approach to
  automatic corpus generation for chinese spelling check}.
\newblock In \emph{Proceedings of the 2018 Conference on Empirical Methods in
  Natural Language Processing, Brussels, Belgium, October 31 - November 4,
  2018}, pages 2517--2527. Association for Computational Linguistics.

\bibitem[{Wu et~al.(2013)Wu, Liu, and Lee}]{DBLP:conf/acl-sighan/WuLL13}
Shih{-}Hung Wu, Chao{-}Lin Liu, and Lung{-}Hao Lee. 2013.
\newblock \href {https://aclanthology.org/W13-4406/} {Chinese spelling check
  evaluation at {SIGHAN} bake-off 2013}.
\newblock In \emph{Proceedings of the Seventh {SIGHAN} Workshop on Chinese
  Language Processing, SIGHAN@IJCNLP 2013, Nagoya, Japan, October 14-18, 2013},
  pages 35--42. Asian Federation of Natural Language Processing.

\bibitem[{Xie et~al.(2015)Xie, Huang, Zhang, Hong, Huang, Chen, and
  Huang}]{DBLP:conf/acl-sighan/XieHZHHCH15}
Weijian Xie, Peijie Huang, Xinrui Zhang, Kaiduo Hong, Qiang Huang, Bingzhou
  Chen, and Lei Huang. 2015.
\newblock \href {https://doi.org/10.18653/v1/W15-3120} {Chinese spelling check
  system based on n-gram model}.
\newblock In \emph{Proceedings of the Eighth {SIGHAN} Workshop on Chinese
  Language Processing, SIGHAN@IJCNLP 2015, Beijing, China, July 30-31, 2015},
  pages 128--136. Association for Computational Linguistics.

\bibitem[{Xu et~al.(2021)Xu, Li, Zhou, Li, Wang, Cao, Huang, and
  Mao}]{DBLP:conf/acl/XuLZLWCHM21}
Heng{-}Da Xu, Zhongli Li, Qingyu Zhou, Chao Li, Zizhen Wang, Yunbo Cao, Heyan
  Huang, and Xian{-}Ling Mao. 2021.
\newblock \href {https://doi.org/10.18653/v1/2021.findings-acl.64} {Read,
  listen, and see: Leveraging multimodal information helps chinese spell
  checking}.
\newblock In \emph{Findings of the Association for Computational Linguistics:
  {ACL/IJCNLP} 2021, Online Event, August 1-6, 2021}, volume {ACL/IJCNLP} 2021
  of \emph{Findings of {ACL}}, pages 716--728. Association for Computational
  Linguistics.

\bibitem[{Yu and Li(2014)}]{DBLP:conf/acl-sighan/YuL14}
Junjie Yu and Zhenghua Li. 2014.
\newblock \href {https://doi.org/10.3115/v1/W14-6835} {Chinese spelling error
  detection and correction based on language model, pronunciation, and shape}.
\newblock In \emph{Proceedings of The Third {CIPS-SIGHAN} Joint Conference on
  Chinese Language Processing, Wuhan, China, October 20-21, 2014}, pages
  220--223. Association for Computational Linguistics.

\bibitem[{Yu et~al.(2014)Yu, Lee, Tseng, and
  Chen}]{DBLP:conf/acl-sighan/YuLTC14}
Liang{-}Chih Yu, Lung{-}Hao Lee, Yuen{-}Hsien Tseng, and Hsin{-}Hsi Chen. 2014.
\newblock \href {https://doi.org/10.3115/v1/W14-6820} {Overview of {SIGHAN}
  2014 bake-off for chinese spelling check}.
\newblock In \emph{Proceedings of The Third {CIPS-SIGHAN} Joint Conference on
  Chinese Language Processing, Wuhan, China, October 20-21, 2014}, pages
  126--132. Association for Computational Linguistics.

\bibitem[{Zhang et~al.(2021)Zhang, Pang, Zhang, Wang, He, Sun, Wu, and
  Wang}]{DBLP:conf/acl/ZhangPZWHSWW21}
Ruiqing Zhang, Chao Pang, Chuanqiang Zhang, Shuohuan Wang, Zhongjun He, Yu~Sun,
  Hua Wu, and Haifeng Wang. 2021.
\newblock \href {https://doi.org/10.18653/v1/2021.findings-acl.198} {Correcting
  chinese spelling errors with phonetic pre-training}.
\newblock In \emph{Findings of the Association for Computational Linguistics:
  {ACL/IJCNLP} 2021, Online Event, August 1-6, 2021}, volume {ACL/IJCNLP} 2021
  of \emph{Findings of {ACL}}, pages 2250--2261. Association for Computational
  Linguistics.

\bibitem[{Zhang et~al.(2020)Zhang, Huang, Liu, and
  Li}]{DBLP:conf/acl/ZhangHLL20}
Shaohua Zhang, Haoran Huang, Jicong Liu, and Hang Li. 2020.
\newblock \href {https://doi.org/10.18653/v1/2020.acl-main.82} {Spelling error
  correction with soft-masked {BERT}}.
\newblock In \emph{Proceedings of the 58th Annual Meeting of the Association
  for Computational Linguistics, {ACL} 2020, Online, July 5-10, 2020}, pages
  882--890. Association for Computational Linguistics.

\end{thebibliography}
\bibliographystyle{acl_natbib}

\appendix

\section{Character-level and Official Evaluation}
\label{sec:other evaluation}
\begin{table*}[t]
\centering\setlength{\tabcolsep}{5pt}
\begin{tabular*}{\textwidth}{@{\extracolsep{\fill}}lccccccc}
\toprule
\multirow{2}{*}{Dataset}  & \multirow{2}{*}{Model}           & \multicolumn{3}{c}{Detection-level}          & \multicolumn{3}{c}{Correction-level}          \\ 
\cmidrule(r){3-5}\cmidrule(r){6-8} 
                            &                                  & D-P           & D-R           & D-F           & C-P           & C-R           & C-F           \\ \midrule
\multirow{4}{*}{SIGHAN15}   & SpellGCN \citep{DBLP:conf/acl/ChengXCJWWCQ20}               & 77.7          & 85.6          & 81.4          & 96.9          & 82.9          & 89.4          \\
                            & PLOME \citep{DBLP:conf/acl/LiuYYZW20}                  & 85.2         & 86.8          & 86.0          & 97.2          & 85.0          & 90.7          \\
                            & CRASpell \citep{liu-etal-2022-craspell}                  &  83.5         & \textbf{89.2}          & 86.3          & 97.1          & \textbf{86.6}          & 91.5          \\
                            & SCOPE (ours)                             & \textbf{86.8}           & 88.9 & \textbf{87.8} & \textbf{97.4} & \textbf{86.6} & \textbf{91.7} \\ 
\bottomrule
% \toprule
% \multirow{3}{*}{SIGHAN14} & PLOME \citep{DBLP:conf/acl/LiuYYZW20}                  & 77.4         & 79.6          & 78.5          & \textbf{98.8}          & 78.8          & 87.7          \\
%                         & CRASpell \citep{liu-etal-2022-craspell}                  &  78.2         & 82.1          & 80.1          & 98.4          & \textbf{80.8}          & \textbf{88.7}          \\
%                         & SCOPE(ours)                             & \textbf{79.9}                        & 81.6          & \textbf{80.7} & 97.8 & 79.8          & 87.9 \\ 
% \bottomrule
% \toprule
% \multirow{3}{*}{SIGHAN13} & PN \citep{DBLP:conf/acl/WangTZ19}             & 56.8          & \textbf{97.4}          & 70.1          & 79.7          & 59.4          & 68.1          \\
%                         & SpellGCN \citep{DBLP:conf/acl/ChengXCJWWCQ20}               & 82.6          & 88.9          & 85.7          & 98.4          & 88.4          & 93.1          \\
%                         & SCOPE(ours)                             & \textbf{92.0}          & 91.0 & \textbf{91.5} & \textbf{98.7}          & \textbf{89.8} & \textbf{94.0} \\ 
% \bottomrule   
\end{tabular*}
\caption{\label{character-level}Character-level performance on the whole test set of SIGHAN15, with baseline results directly taken from their respective literatures.}
\end{table*}

\begin{table*}[t]
\centering
\begin{tabular*}{\textwidth}{@{\extracolsep{\fill}}lccccccc}
\toprule
\multirow{2}{*}{Dataset}  & \multirow{2}{*}{Model} & \multicolumn{3}{c}{Detection-level}          & \multicolumn{3}{c}{Correction-level}          \\ 
\cmidrule(r){3-5}\cmidrule(r){6-8}  
                            &                        & D-P           & D-R           & D-F           & C-P           & C-R           & C-F           \\ \midrule
\multirow{4}{*}{SIGHAN15} & SpellGCN \citep{DBLP:conf/acl/ChengXCJWWCQ20}     & 85.9          & 80.6          & 83.1          & 85.4          & 77.6          & 81.3          \\
& PLOME \citep{DBLP:conf/acl/LiuYYZW20}        & 87.9          & 80.9          & 84.3          & 87.6          & 78.3          & 82.7          \\
                            & GAD \citep{DBLP:conf/acl/GuoNWZX21}                  & 86.0         & 80.4          & 83.1          & 85.6          & 77.8          & 81.5          \\
                            & SpellBERT \citep{DBLP:conf/emnlp/JiYQ21}        & 87.5          & 73.6          & 80.0          & 87.1          & 71.5          & 78.5          \\
                            & SCOPE (ours)                   & \textbf{89.4} & \textbf{84.3} & \textbf{86.3} & \textbf{89.2} & \textbf{82.4} & \textbf{85.7} \\ 
\bottomrule
% \toprule
% \multirow{3}{*}{SIGHAN14} & SpellGCN \citep{DBLP:conf/acl/ChengXCJWWCQ20}     & 83.1          & 69.5          & 75.7          & 82.8          & 67.8          & 74.5          \\
%                             & SpellBERT \citep{DBLP:conf/emnlp/JiYQ21}  & 83.1          & 62.0          & 71.0          & 82.9          & 61.2          & 70.4          \\
%                             & SCOPE(ours)                   & \textbf{83.9} & \textbf{73.1} & \textbf{78.1} & \textbf{83.6} & \textbf{71.5} & \textbf{77.1} \\ 
% \bottomrule
\end{tabular*}
\caption{\label{official tool}Official evaluation results on the whole test set of SIGHAN15, with baseline results directly taken from their respective literatures.}
\end{table*}

This section further compares SCOPE to some recently proposed methods that have not been evaluated with sentence-level metrics, but instead with character-level and/or official evaluation metrics. These baseline methods include:
\begin{itemize}
    \item {\it SpellBERT} \citep{DBLP:conf/emnlp/JiYQ21} uses a lightweight pre-trained model for CSC, encoding phonetic and visual features with GNNs. 
    \item {\it GAD} \citep{DBLP:conf/acl/GuoNWZX21} models the global dependency between all candidate characters by a global attention decoder. 
    \item {\it CRASpell} \citep{liu-etal-2022-craspell} constructs a noise modeling module that makes their model robust to consecutive spelling errors, with a copy mechanism to handle over-correction. 
\end{itemize}
For reference, we also include two previously compared baselines SpellGCN \cite{DBLP:conf/acl/ChengXCJWWCQ20} and PLOME \citep{DBLP:conf/acl/LiuYYZW20} that have their results reported on these new metrics. We use the code released by REALISE \citep{DBLP:conf/acl/XuLZLWCHM21}\footnote{\url{https://github.com/DaDaMrX/ReaLiSe}} for sentence-level evaluation and the code released by CRASpell \citep{liu-etal-2022-craspell}\footnote{\url{https://github.com/liushulinle/CRASpell}} for character-level evaluation. The official evaluation scripts are provided along with the datasets.\footnote{\url{http://ir.itc.ntnu.edu.tw/lre/sighan7csc.html}}$^,$\footnote{\url{http://ir.itc.ntnu.edu.tw/lre/clp14csc.html}}$^,$\footnote{\url{http://ir.itc.ntnu.edu.tw/lre/sighan8csc.html}}

The results are shown in Table \ref{character-level} and Table \ref{official tool}. 
We can see that regardless of the evaluation scenarios, SCOPE consistently outperforms all the baselines in almost all metrics, verifying its effectiveness and superiority for CSC. 

\section{Hyperparameter Search}
We conduct a hyperparameter search for learning rate, batch size and epoch. Learning rate is tuned from $\{2\!\times\! 10^{-5}, 5\!\times\! 10^{-5}\}$, batch size from $\{48, 64,$ $96\}$ and epoch from $\{20, 30\}$. There are 12 hyperparameter search trials in total on each dataset. The optimal configurations are given in Section~\ref{subsec:setups}.

\end{CJK}

\end{document}